\documentclass[german,english]{bvm2019}

\title{Feasibility of Colon Cancer Detection in Confocal Laser Microscopy Images Using Convolution Neural Networks}

\titlerunning{Colon Cancer Detection in CLM Images Using CNNs}

\subtitle{}
\usepackage[multi-part-units=single]{siunitx}
\usepackage{graphicx}
\usepackage[outdir=./]{epstopdf}
\usepackage{bm}
\author{Nils Gessert$^{1}$, Lukas Wittig$^{2}$, Daniel Dr\"omann$^{2}$, Tobias Keck$^3$, Alexander Schlaefer$^{1}$, David B. Ellebrecht$^{3}$}

\authorrunning{Gessert et al.}

\institute{%
$^1$Institute of Medical Technology, Hamburg University of Technology\\
$^2$Department of Pulmology, University Medical Centre Schleswig-Holstein\\
$^3$Department of Surgery, University Medical Centre Schleswig-Holstein\\
}
\email{nils.gessert@tuhh.de}

\begin{document}

%
\selectlanguage{english}

\maketitle

\begin{abstract}

Histological evaluation of tissue samples is a typical approach to identify colorectal cancer metastases in the peritoneum. For immediate assessment, reliable and real-time in-vivo imaging would be required. For example, intraoperative confocal laser microscopy has been shown to be suitable for distinguishing organs and also malignant and benign tissue. So far, the analysis is done by human experts. We investigate the feasibility of automatic colon cancer classification from confocal laser microscopy images using deep learning models. We overcome very small dataset sizes through transfer learning with state-of-the-art architectures. We achieve an accuracy of $\SI{89.1}{\percent}$ for cancer detection in the peritoneum which indicates viability as an intraoperative decision support system.


\end{abstract}

\section{Introduction}

Colorectal cancer is one of the most common types of cancer \cite{2655-01}. Due to metastatic spread, peritoneal carcinomatosis can occur in later stages which often leads to substantially shorter survival times \cite{2655-02}.  Therefore, reliable detection of metastases is important. Typical imaging modalities such as magnetic resonance imaging and computed tomography currently lack the required resolution and intraoperative availability. Therefore, an intraoperative device using confocal laser microscopy (CLM) has been proposed \cite{2655-03} which offers submicrometer resolution. 

In the above-mentioned study, colon carcinoma cells were implanted into the colon and peritoneum of ten rats. After seven days of tumor growth, laparotomy was carried out for subsequent in-vivo CLM. For each subject, healthy colon tissue, malignant colon tissue, healthy peritoneum and malignant peritoneum were scanned. The study showed that different organs, as well as malignant and non-malignant regions could be distinguished by experts. 

To further improve the intraoperative assessment by CLM, image processing methods can be used for automatic and fast tissue characterization. Recently, deep learning methods have shown remarkable success for a variety of medical segmentation and classification tasks \cite{2655-04} where human-level performance was achieved \cite{2655-05}.

We investigate the feasibility of deep learning-based colon cancer detection from CLM images. We consider several classification problems with the four classes "colon normal", "colon malignant", "peritoneum normal" and "peritoneum malignant". In particular, we investigate both the differentiability of organs and also of malignant and non-malignant tissue both for the colon and peritoneum. As we are dealing with a very small dataset we employ transfer learning which has been shown to improve performance for a variety of medical learning problems \cite{2655-06,2655-07}. We use the state-of-the-art models Densenet121 \cite{2655-08} and SE-Resnext50 \cite{2655-09} which are pretrained on the ImageNet dataset.

\section{Methods}

\subsection{Dataset}

\begin{figure}
\centering
\includegraphics[width=1.0\textwidth]{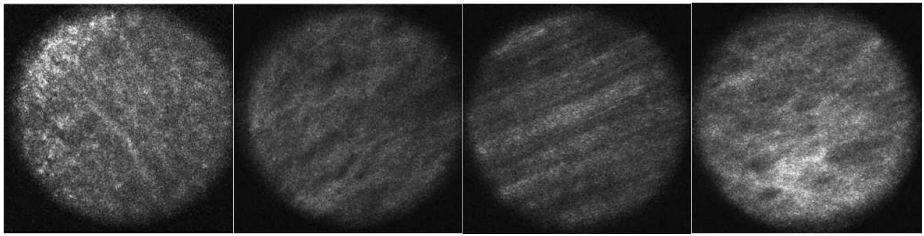}
\caption{Examples of the four different classes. From left to right, healthy colon tissue, malignant colon tissue, healthy peritoneum tissue and malignant peritoneum tissue.}
\label{fig:examples}
\end{figure}

The dataset we use was kindly provided to us by the authors of a previous study on CLM \cite{2655-03}. The dataset was acquired at the University Hospital Schleswig-Holstein in L\"ubeck using a custom intraoperative CLM device. The CLM device (Karl Storz GmbH \& Co KG, Tuttlingen, Germany) covers a field of view of $\SI{300}{\micro\metre} \times \SI{300}{\micro\metre}$ with a resolution of $384\times 384$ pixels. The images were obtained from ten rats where colon adenocarcinoma cells had been implanted into the colon and peritoneum seven days before scanning. For each subject, images of healthy colon tissue (HC), malignant colon tissue (MC), healthy peritoneum tissue (HP) and malignant peritoneum tissue (MP) were obtained. In total, there are 533 images of class HC, 309 images of class MC, 343 images of class HP and 392 images of class MP which results in a total dataset size of 1577 images. Note, that for one subject there are no images of class HC and for one subject there are no images of class MP. Example cases for each class are shown in Figure~\ref{fig:examples}. The assignment of classes for each image was performed based on subsequent histological evaluation of resected tissue from the scanning area. 

We split the dataset in a leave-one-subject-out cross-validation scheme, i.e., we consider ten different dataset splits where images from one subject are left out for evaluation. If a required class is missing, the subject's validation split is omitted. We consider three classification problems in total. First, we address the binary classification task HC versus HP which provides information on whether the organs can be differentiated in principle. Next, we consider the learning problems HC versus MC and HP versus MP which investigates the feasibility of detecting malignant tissue from CLM images. 

\subsection{Models and Training} \label{sec:models}

\begin{figure}
\centering
\includegraphics[angle=90,width=1.0\textwidth]{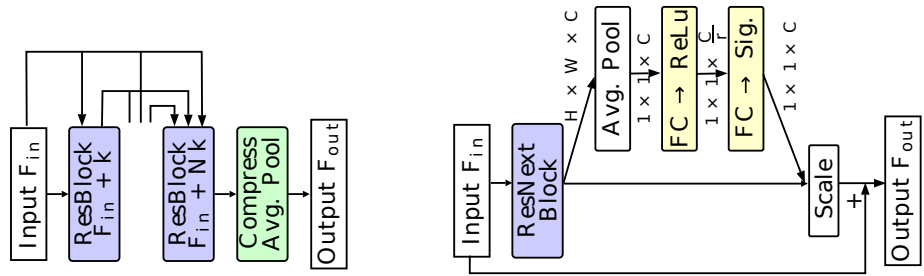}
\caption{The key concepts of the architecture we employ. The shown modules replace sets of standard convolutional layers in the architecture. Left, a Densenet \cite{2655-08} block is shown. Right, an SE block is shown for the Resnext architecture \cite{2655-09}.}
\label{fig:models}
\end{figure}

We employ convolutional neural networks (CNNs) for the classification tasks at hand. The images are directly fed into a CNN which learns to extract relevant features and also perform classification at its output. We employ the two state-of-the-art architectures Densenet121 \cite{2655-08} and SE-Resnext50 \cite{2655-09}. Densenet121 follows the principle of densely connected layers, i.e., features computed within a convolutional layers are also reused in subsequent layers. In this way, the architecture is very efficient in terms of the number of learnable parameters as features are reused heavily. Considering the small dataset size at hand, this can be very beneficial. The SE-Resnext50 architecture is based on the Resnext principle \cite{2655-10} where feature extraction is performed by multiple, parallel paths. In addition, squeeze and excitation (SE) modules are incorporated into the model which perform a feature recalibration step. In standard convolutions the aggregation of features is learned implicitly through a summation. Instead, the SE modules explicitly model dependencies between learned features which increases the models' representational power. The building blocks of the two concepts are shown in Figure~\ref{fig:models}.

To overcome the general lack of data, we use transfer learning, i.e. the models are pretrained on the ImageNet dataset. During training we fine tune all weights. For comparison, we also consider training from scratch. The pretrained models' input layer contains three channels. We put the gray-scale CLM images into one channel and set the other channels to zero. We cut off the last layer and add fully-connected layer with two outputs for binary classification.

During training, we use online data augmentation with unscaled random crops of size $224\times 224$ from the original images of size $384\times 384$. Also, we use random flipping along both dimensions and random changes in brightness and contrast. For stochastic gradient descent we employ Adam with a batch size of $40$ and learning rate of $\num{1e-5}$ and we train for $125$ epochs. For evaluation, we use multi-crop evaluation with $N_c = 9$ crops. The predictions of all crops are averaged into a final prediction for each image. The models are implemented in PyTorch.

\section{Results}

\begin{table}[t]
\caption{The results of all our deep learning experiments. The mean values for leave-one-subject-out cross-validation are shown. Dense refers to the Densenet121 model, SE-RX refers to the SE-Resnext50 model. TL refers to transfer learning and SRC refers to training from scratch. For each training scenario, the best performing value is marked bold. All values are given in percent. The sensitivity is given with respect to the cancer class and for the case of organ differentiation it is given with respect to the peritoneum class.}
\label{tab:res}
\begin{tabular*}{\textwidth}{l@{\extracolsep\fill}lllll}
\hline
 & & Accuracy & Sensitivity & Specificity & F1-Score \\
\hline
\parbox[t]{2mm}{\multirow{4}{*}{\rotatebox[origin=c]{90}{HC vs. HP}}} & Dense TL & $\bm{90.8}$ & $\bm{80.2}$ & $\bm{93.9}$ & $\bm{91.7}$ \\
& Dense SRC & $78.5$ & $74.2$ & $78.1$ & $79.1$ \\
& SE-RX TL & $89.3$ & $78.6$ & $90.3$ & $90.5$ \\
& SE-RX SRC  & $70.8$ & $77.9$ & $67.3$ & $72.6$ \\
\hline
\parbox[t]{2mm}{\multirow{4}{*}{\rotatebox[origin=c]{90}{HC vs. MC}}} & Dense TL & $\bm{66.7}$ & $74.1$ & $64.8$ & $\bm{69.0}$ \\
& Dense SRC & $60.0$ & $\bm{81.0}$ & $50.7$ & $63.6$ \\
& SE-RX TL & $58.9$ & $69.8$ & $57.3$ & $62.6$  \\
& SE-RX SRC & $64.5$ & $69.5$ & $\bm{67.1}$ & $65.6$  \\
 \hline
\parbox[t]{2mm}{\multirow{4}{*}{\rotatebox[origin=c]{90}{HP vs. MP}}} & Dense TL & $\bm{89.1}$ & $80.9$ & $\bm{87.2}$ & $\bm{90.0}$ \\
& Dense SRC & $77.0$ & $70.8$ & $70.2$ & $79.3$ \\
& SE-RX TL & $83.2$ & $72.5$ & $86.9$ & $84.9$  \\
& SE-RX SRC & $77.3$ & $\bm{85.4}$ & $64.7$ & $77.6$  \\
\hline
\end{tabular*}
\end{table}

All results are shown in Table~\ref{tab:res}. In terms of metrics, we report accuracy, sensitivity, specificity and the F1-score. For each of the three training scenarios, HC versus HP, HC versus MC and HP versus MP, we consider the architectures described in Section~\ref{sec:models}. Also, for each case we consider training from scratch and fine-tuning after pretraining on ImageNet. In general, the classification accuracy is high for the distinction of organs and also the differentiation between benign and malignant tissue of the peritoneum. However, the performance for cancer detection in the colon is significantly lower. Comparing the two architectures, the performance is very similar with Densenet121 generally performing slightly better. Using transfer learning with pretrained architectures improves performance substantially for most cases. 

\section{Discussion}

In this study we investigate the feasibility of detecting colon cancer from confocal laser microscopy (CLM) images using deep learning models. This extends a previous study where the feasibility of cancer detection from CLM images by experts was shown \cite{2655-03}. Here, we use two state-of-the-art deep learning architectures to automatically detect cancer from CLM images.
As a baseline, we consider the task of differentiating healthy tissue from the colon and the peritoneum. With an F1-score of $91.7$, the best model, Densenet121, shows a high performance which indicates that different organs can be well distinguished in CLM images by deep learning models. It is notable that without pretraining performance drops substantially across all metrics. This highlights the effectiveness of transfer learning for a particularly small dataset \cite{2655-06}.
Regarding the detection of malignant tissue in the peritoneum, the model performance is also very high with Densenet121 performing best. It is notable that Densenet121 generally performs better than SE-Resnext50 in our study while the latter clearly outperforms the former on the ImageNet dataset \cite{2655-09}. This is likely tied to Densenet121 having significantly fewer parameters which prevents overfitting with the small dataset. Also, the performance difference between training from scratch and transfer learning is larger for Densenet121. This indicates, that Densenet121 benefits more from the pretrained weights.
Considering the detection of malignant tissue in the colon, the performance is significantly lower compared to the other tasks. It should be noted that the performance difference is most obvious in the specificity. Thus, most cases of cancer are detected but a lot of false positives occur as well. This might be tied to the heterogeneous appearance of the colon in different areas which makes the learning task very challenging due to the small dataset size. Also, carcinoma cells transform from healthy tissue via adenoma to carcinoma. Thus, healthy and malignant tissue can have a similar appearance which might complicate the learning problem.

Overall, we showed that automatic organ differentiation and cancer detection from CLM images is feasible using pretrained convolutional neural networks. For future work, more data could be acquired and the detection of malignant tissue in the colon area could be studied further.

\bibliographystyle{bvm2019}

\bibliography{2655}

\begin{thebibliography}{10}

\bibitem{2655-01}
Torre LA, Bray F, Siegel RL, Ferlay J, Lortet-Tieulent J, Jemal A.
\newblock Global cancer statistics, 2012.
\newblock CA: A Cancer Journal for Clinicians. 2015;65(2):87--108.

\bibitem{2655-02}
Franko J, Shi Q, Goldman CD, Pockaj BA, Nelson GD, Goldberg RM, et~al.
\newblock Treatment of colorectal peritoneal carcinomatosis with systemic
  chemotherapy: a pooled analysis of north central cancer treatment group phase
  III trials N9741 and N9841.
\newblock Journal of Clinical Oncology. 2012;30(3):263.

\bibitem{2655-03}
Ellebrecht DB, Kuempers C, Horn M, Keck T, Kleemann M.
\newblock Confocal laser microscopy as novel approach for real-time and in-vivo
  tissue examination during minimal-invasive surgery in colon cancer.
\newblock Surgical Endoscopy. 2018; p. 1--7.

\bibitem{2655-04}
Litjens G, Kooi T, Bejnordi BE, Setio AAA, Ciompi F, Ghafoorian M, et~al.
\newblock A survey on deep learning in medical image analysis.
\newblock Medical Image Analysis. 2017;42:60--88.

\bibitem{2655-05}
Esteva A, Kuprel B, Novoa RA, Ko J, Swetter SM, Blau HM, et~al.
\newblock Dermatologist-level classification of skin cancer with deep neural
  networks.
\newblock Nature. 2017;542(7639):115.

\bibitem{2655-06}
Hoo-Chang S, Roth HR, Gao M, Lu L, Xu Z, Nogues I, et~al.
\newblock Deep convolutional neural networks for computer-aided detection: CNN
  architectures, dataset characteristics and transfer learning.
\newblock IEEE Transactions on Medical Imaging. 2016;35(5):1285.

\bibitem{2655-07}
Gessert N, Lutz M, Heyder M, Latus S, Leistner DM, Abdelwahed YS, et~al.
\newblock Automatic Plaque Detection in IVOCT Pullbacks Using Convolutional
  Neural Networks.
\newblock IEEE Transactions on Medical Imaging. 2018; p. 1--9.

\bibitem{2655-08}
Huang G, Liu Z, Weinberger KQ, van~der Maaten L.
\newblock Densely connected convolutional networks.
\newblock In: Proceedings of the IEEE CVPR; 2017. .

\bibitem{2655-09}
Hu J, Shen L, Sun G.
\newblock Squeeze-and-Excitation Networks.
\newblock In: Proceedings of the IEEE CVPR; 2018. .

\bibitem{2655-10}
Xie S, Girshick R, Doll{\'a}r P, Tu Z, He K.
\newblock Aggregated residual transformations for deep neural networks.
\newblock In: Proceedings of the IEEE CVPR. IEEE; 2017.  p. 5987--5995.

\end{thebibliography}
\end{document}